\documentclass[]{article}
\usepackage[letterpaper]{geometry}
\usepackage{amta2022}
\usepackage{MnSymbol}
\usepackage{times}
\usepackage{url}
\usepackage{latexsym}
\usepackage{natbib}
\usepackage{layout}

\usepackage{upgreek}

\usepackage{enumitem}

\usepackage{amsfonts}

\usepackage{bm}

\usepackage{amsmath}

\usepackage{xurl}

\usepackage{tabularx}
\usepackage{booktabs}% http://ctan.org/pkg/booktabs
\newcolumntype{R}{>{\centering\arraybackslash}X}

\usepackage{makecell}

\usepackage{adjustbox}
\newcolumntype{Y}{>{\centering\arraybackslash}X}

\newcolumntype{L}{>{\centering\arraybackslash}m{3cm}}

\usepackage{multirow}

\newcolumntype{?}{!{\vrule width 1.5pt}}

\usepackage{times}
\usepackage{textcomp}

\usepackage{blindtext}
\usepackage{graphicx}
\usepackage{lettrine}

\usepackage{varwidth}

\usepackage{tabu}

\usepackage[lofdepth,lotdepth]{subfig}

\usepackage{url}

\usepackage{float}
\usepackage{caption}
\usepackage{multirow}
\usepackage{hyperref}

\usepackage{xparse}
\newcommand\numberthis{\addtocounter{equation}{1}\tag{\theequation}}

\DeclareMathOperator*{\argmin}{arg\,min}

\begin{document}

\title{\bf Fast Vocabulary Projection Method via Clustering for Multilingual Machine Translation on GPU}

\author{
    \name{\bf Hossam Amer} \hfill
        \addr{\texttt{hossamamer [at] microsoft.com}} \\
    \name{\bf Young Jin Kim} \hfill
       \addr{\texttt{youki [at] microsoft.com}}  \\
    \name{\bf Mohamed Afify} \hfill
        \addr{\texttt{mafify [at] microsoft.com}} \\
    \name{\bf Hitokazu Matsushita} \hfill
        \addr{\texttt{hitokazu.matsushita [at] microsoft.com}} \\
    \name{\bf Hany Hassan Awadallah} \hfill
        \addr{\texttt{hanyh [at] microsoft.com}} \\
      \addr{Microsoft}
}
\maketitle
\pagestyle{empty}

% %%%%%%%%%%%%%%%%%%%%%%%%%%%

\begin{abstract}

Multilingual Neural Machine Translation has been showing great success using transformer models. Deploying these models is challenging because they usually require large vocabulary (vocab) sizes for various languages. This limits the speed of predicting the output tokens in the last vocab projection layer. To alleviate these challenges, this paper proposes a fast vocabulary projection method via clustering which can be used for multilingual transformers on GPUs. First, we offline split the vocab search space into disjoint clusters given the hidden context vector of the decoder output, which results in much smaller vocab columns for vocab projection. Second, at inference time, the proposed method predicts the clusters and candidate active tokens for hidden context vectors at the vocab projection. This paper also includes analysis of different ways of building these clusters in multilingual settings. Our results show end-to-end speed gains in float16 GPU inference up to 25\% while maintaining the BLEU score and slightly increasing memory cost. The proposed method speeds up the vocab projection step itself by up to 2.6x. We also conduct an extensive human evaluation to verify the proposed method preserves the quality of the translations from the original model.
\end{abstract}

%%%%%%%%%%%%%%%%%%%%%%%%%%

\section{Introduction}
\label{Introduction}
Neural machine translation (NMT) has witnessed significant advances by the introduction of the transformer model \citep{vaswani2017attention} where excellent performance has been shown for bilingual translation initially, mainly to and from English. Later, the model has been extended to multiple language pairs e.g. \citep{johnson-etal-2017-googles} and is referred to as multilingual neural machine translation (MNMT). The multilingual model usually comes with a significant increase in the number of parameters and the vocabulary (vocab) size to accommodate various languages.

This work focuses on implementing efficient vocab projection in transformer models. The problem becomes very important for MNMT. While bilingual NMT typically use vocab of around 32K sub-words, MNMT has significantly larger vocab size to adequately cover all the languages.
For example, the recent MNMT submission from Microsoft to the WMT21 shared task that covers 100+ languages has a vocab size of 250K sub-words \citep{Yang2021MSWMT21}. The increased vocab makes the projection very compute-intensive and it can take up to 25\% of the total computation in our internal benchmark. Hence, the  need to speedup this operation becomes even more important.

Fast vocab projection is well-studied for NMT and natural language processing (NLP). In \citep{shi-knight-2017-speeding}, the authors propose two methods to reduce the effective vocab for NMT. The first formulates the problem as a nearest neighbor search and uses locality sensitive hashing (LSH) to speedup the computation while the second uses alignment information to select a subset of the vocab based on the input. It is found that the alignment-based method leads to around 2X speedup. The work \citep{chen2018learning} proposes an efficient screening model to exploit the clustering structure of context features right before vocab projection. The authors formulate a joint optimization problem to learn the clusters and their corresponding vocab subspace. In the same paper, a thorough comparison to the existing literature is done including graph-based nearest-neighbor search \citep{Zhang2018FGD}, SVD approximation\citep{Shim2017SVDSoftmaxFS}, a modified version of hierarchical softmax \citep{Grave2017adaptivesoftmax} and locality sensitive hashing (LSH) for maximum inner product search (MIPS) \citep{Neyshabur2014LSH}. It is shown that the clustering solution can outperform the latter techniques in terms of accuracy-speed trade-off in bilingual LSTMs on CPU.

This paper extends the clustering based approach from \citep{chen2018learning} to MNMT for both dense and sparse transformer models on GPU. To simplify integration and accessibility in the MNMT large-scale model settings, we propose to use kmeans clustering to split the vocab search space into disjoint clusters given the hidden context vector of the decoder output. The multilingual extension comprises experimenting with different ways to build the cluster maps as well as testing different configurations such as number of centroids and corresponding vocab subspace to make the method efficiently work at large scale. In \citep{Shi2018LSHGPU}, it is shown how to extend LSH to run efficiently within beam search on GPU but here we use and develop optimized kernels. These kernels are developed for FasterTransformer, a highly optimized transformer-based encoder and decoder implementation offered by NVIDIA\footnote{\url{https://github.com/NVIDIA/FasterTransformer}}. Serving GPU inference is essential because GPUs give several orders of magnitude speedups relative to CPUs for large transformer models \footnote{\url{https://www.nvidia.com/en-us/on-demand/session/gtcspring22-s42518/}}. Experimental results show end-to-end speed gains in float16 GPU inference up to 25\% while maintaining the BLEU score and slightly increasing memory cost. The proposed method speeds up the vocab projection step itself by up to 2.6x. We also conduct an extensive human evaluation to verify the proposed method preserves the quality of the translations from the original model.

The paper is organized as follows. Section \ref{Method} describes the proposed method focusing on the multilingual aspect and GPU implementation. This is followed by experimental results for both dense and sparse transformer models and for 6 translation directions in Section \ref{experiments}. Finally, Section \ref{conclusion} concludes the paper.

\section{Method}
\label{Method}

\subsection{Problem Motivation}
\label{sec:motivation}

The last layer of decoder in MNMT transformer models is typically a vocab projection layer followed by softmax activation to get the predicted probability of output tokens. Let $W \in \mathbb{R}^{d \times N}$ be a weight matrix of the trained model, $b \in \mathbb{R}^{N}$ be the bias of the trained model, and $h \in \mathbb{R}^{d \times M}$ be the hidden context vector before the vocab projection layer. Here, $N$ is the vocab size of the MNMT transformer model, $d$ is the transformer embedding dimension, and $M$ is the product of batch and beam sizes. For a total of $M$ tokens and current time step $t$, we compute the vocab projection as follows:

\begin{equation}
    z_{(m, t)} = W^{T}h_{(m,t)} + b
\label{eqn:vocab_proj}
\end{equation}

\noindent where $z_{(m, t)}$ is the logits of the current input token, $m$, at the current time step $t$. To compute the probabilities of the predicted token $\hat{p}_{{m,t}}$, we compute $\hat{p}_{{m,t}} = softmax(z_{(m, t)})$. These probabilities are sorted in descending order for which they are used for the search algorithms such as beam search, greedy search, sampling, and etc. in machine translation.

In MNMT transformer models, the weights for vocab projection , $W$, is usually large to cover different languages. Our profiling results show that vocab projection occupies about 25\% or more depending on the beam and batch size from the end-to-end inference time. To speedup the inference while keeping the accuracy under control, we propose a vocab projection method via clustering for MNMT transformer models. The proposed vocab projection
consists of two major steps: 1) Offline training of clusters based on hidden context vectors which is the output of the last decoder layer; 2) Online inference using the created clusters. Both steps will be explained in the following sections.

\subsection{Offline Training of Clusters}
\label{sec:cluster_training}

\begin{figure} [htbp]
      \centering
             \includegraphics[width=1\linewidth]{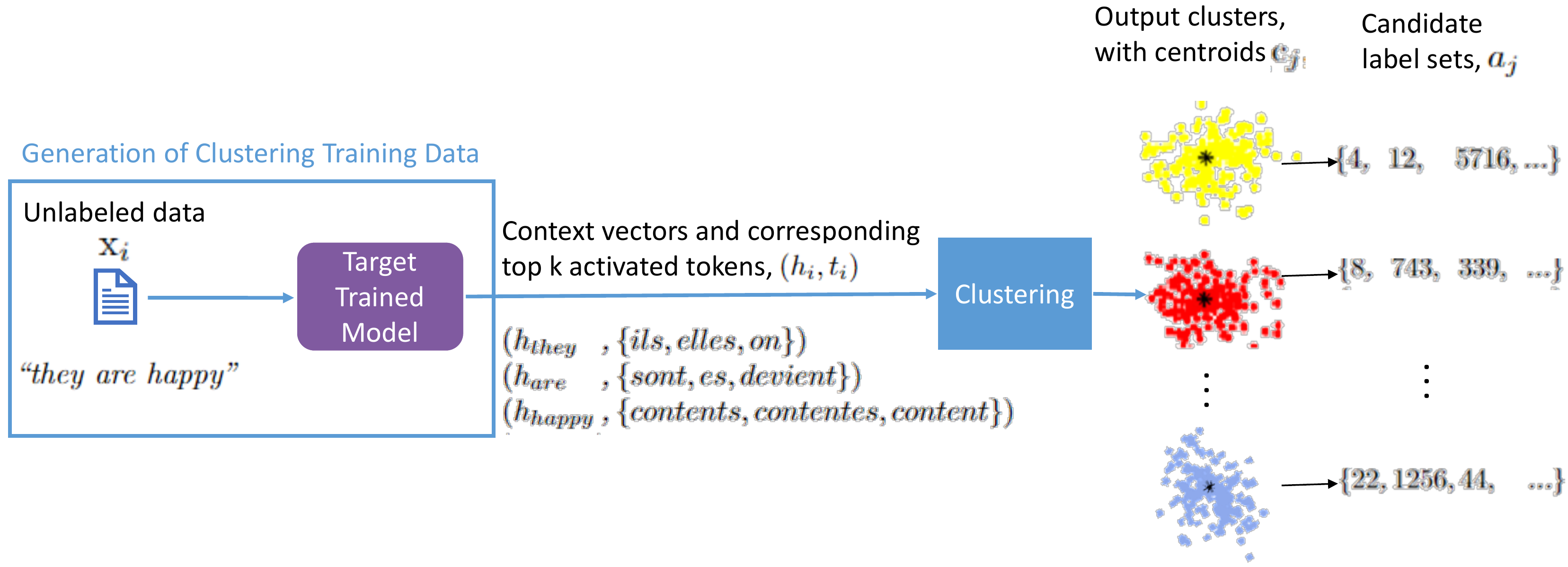}
      \caption{Overview of Vocab Projection via Clustering.}
      \label{fig:clustering_overview}
\end{figure}

Figure \ref{fig:clustering_overview} shows an overview of the offline training step of the proposed clustering-based vocab projection method. Using an unlabelled training set, this method first records the hidden context vectors $h_{(m,t)}$ and corresponding K candidate likely tokens from $\hat{p}_{{m,t}}$ given the pre-trained model. K here indicates the number of likely predicted output tokens for each input token. The use of unlabelled data is a benefit of the proposed method as data is not always available in MNMT. Next, we partition the context vectors into disjoint clusters, $C$, of a specific number of centroids, where similar context vectors are grouped in the same cluster.  From the context vector members of each cluster, we construct an active tokens label set called $A$. Each active token set, $aj \in A$, of the corresponding cluster is the union of the K predicted tokens of all members of this cluster. For example at K=3, suppose that we have only 2 tokens from the training set in the same cluster has the following predicted tokens $\{2, 4, 6\}, \{2, 8, 9\}$ according to the model's predicted probabilities. Then, the $a_j$ for this cluster will be $\{2, 4, 6, 8, 9\}$.

\begin{figure} [htbp]
      \centering
             \includegraphics[width=0.7\linewidth]{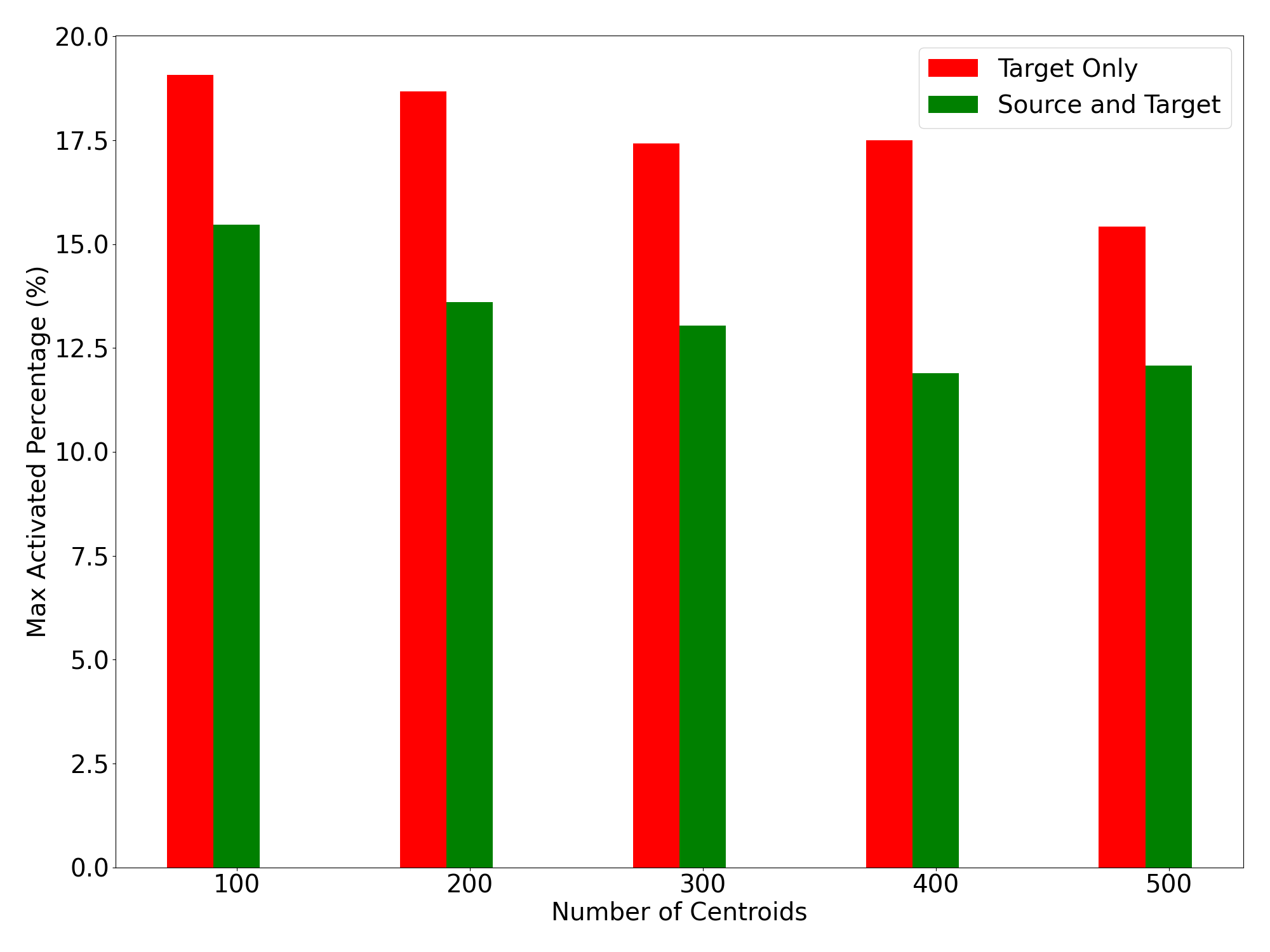}
      \caption{Comparison Between Ways to Build the Clusters and Active Tokens.}
      \label{fig:multilingual_src_and_target_clustering}
\end{figure}

There are different ways to create the clusters and corresponding active tokens set to utilize different information such as source and target languages. Among these ways, we can create the clusters as well as the active tokens while assuming either the target language is only known or the source and target language are known. To compare these two ways, we carried out an experiment on the Italian-to-English (ItEn) language direction, where we varied the number of clusters from 100 to 500 with a step size of 100 at K=1. To construct the clusters of ItEn using source and target information, we feed ItEn training data into the model and get the corresponding clusters and active tokens. For target "en" only known, we feed ItEn, French-to-English (FrEn), and Spanish-to-English (EsEn), data to the model. Later, we  run clustering with this mixture of data and construct the corresponding active tokens. In both cases, we measured the maximum percentage of active vocab columns when the number of clusters ranges from 100 to 500. For instance, Figure \ref{fig:multilingual_src_and_target_clustering} shows that the percentage of maximum activated vocabs among whole vocabs with 400 clusters is 17.5\% when the target only is known, while the percentage is 12.5\% when both source and target information is used. Because we favor speed improvements, we chose to run the rest of all our experiments by constructing the models while assuming that source and target languages are known, which is typical in machine translation. This results in fewer activated vocabs for the projection and a faster matrix product at the expense of only an insignificant increase in memory costs as will be shown in the results section.

\subsection{Vocab Projection via Clustering for GPU Inference}

This section outlines the proposed method steps at inference time. The input of our method is the number of clusters, corresponding centroid vectors and active tokens set obtained from offline training and hidden context vectors. To compute the output logits, the first step of vocab projection via clustering at inference time is to adaptively get the predicted selected cluster indices based on the hidden context vectors. Suppose that $G$ is the cluster indices of the hidden context vectors, and $C_j$ is the centroid vector of cluster $j$, we compute the predicted centroid indices using the Euclidean distance for a token $m$ at time step $t$ as follows:

\begin{align*}
  G_{(m,t)} = & \argmin_{j} \|C_{j}^2 - 2h_{(m,t)}^{test}C_{j}\|
    \numberthis \label{eq:centroid_selection}
\end{align*}

\noindent where $j$ is between 0 and a given number of clusters. Since the centroid vectors, $C_j$ are constant at inference time, we pre-compute $C_j^2$ for all $j$ and save $C_j^2$ in GPU memory.

In practice, the inference is usually done in batch mode where the batch of context vectors can be of different nature and can belong to different clusters as well as candidate label sets. A possible approach is to independently calculate vocab projection for each context vector based on its cluster and corresponding candidate label set. However, this computation will limit the speed on GPUs because they by architecture typically prefer large-scale data in a single computation to hide latency \citep{cheng2014professional}. To overcome this challenge, we propose to union the candidate label sets of the context vectors motivated by the possible overlap of label candidate sets within the beam. Although the vocab projection weight matrix of this union set will be larger than the one for independent computation, a single kernel GPU launch for the batch saves inference time while maintaining accuracy.

To efficiently compute this union on GPU, we create a boolean list to indicate whether a vocab token should be active (=1), which is initialized as inactive (=0). We then launch threads to turn on particular vocab columns in this boolean list depending on the active tokens corresponding to the predicted centroids. Suppose that the predicted centroids list is $G=\{0,1,2\}$ and corresponding active tokens set is $\{2, 4, 6\}, \{2, 8, 9\}, \{1, 3\}$, respectively. The union of the latter token list yields a reduced vocab column set: $\{1, 2, 3, 4, 6, 8, 9\}$. To get the union, we construct a boolean list equals to [0, 1, 1, 1, 1, 0, 1, 0, 1, 1]. Using the CUB Toolkit function on GPU by Nvidea routines, we efficiently get the vocab columns where this boolean list equals 1. The final reduced weight matrix is substituted in Equation \ref{eqn:vocab_proj} using the newly released gather and scatter fusion with GEMM kernel from Nvidia CUTLASS 2.9. To put together our method steps at inference time, we list them as follows:

\begin{enumerate}[noitemsep,nolistsep]
    \item Compute the predicted centroids $c$ by substituting the context vectors in Equation \ref{eq:centroid_selection}.
    \item Get the activated weight column index of the reduced matrix from the union of active tokens for the given predicted centroids.
    \item Compute the logits by substituting the reduced weight matrix in Equation \ref{eqn:vocab_proj}.
    \item Scatter the computed logits in their original locations while setting the non-included weight positions to -INF.
    \item Compute the softmax based on this scattered logits matrix.
\end{enumerate}

\noindent \textbf{Theoretical speed-up} Based on the steps above, the proposed method turns the theoretical complexity of the vocab projection in Equation \ref{eqn:vocab_proj} from $O(dN)$ to $O((r+\bar{N})d)$. The average number of activated weight columns in the reduced vocab matrix, $\bar{N}$, is much less than $N$. $\bar{N}$ is at most to $\approx$15\% of $N$ as shown in the results section. Also, the number of centroids, $r$ is far less than $N$ (typically set to 2000).

%%%%%%%%%%%%%%%%%%%%%%%%%%

\section{Experimental Results}
\label{experiments}

In this section, we present the evaluation results for the proposed vocab projection method in a multi-lingual transformer setting on GPU. We first introduce the experimental setup, then
assess the impact of the proposed method on vocab projection itself as well as show the speed and accuracy performance results. We conducted an extensive human evaluation to verify the translation quality is preserved. Finally, we study how different configurations such as the number of centroids and K likely tokens affect the performance of our method.

\subsection{Experimental Setup}

\textbf{Task and Models} We employ the proposed method in MNMT via two transformer models: 1) ZCode Dense; 2) ZCode M3 reviewed and presented in \cite{kim2021scalable}. Both models are based on the transformer encoder-decoder architecture \citep{vaswani2017attention}. For better inference efficiency, the architecture of these models utilizes the deeper encoders and shallower decoders architecture presented in 
\citep{kim2019research} and \citep{kasai2020deep}. Also, both use pre-layer normalization which is known to be more stable for the deeper transformer architecture presented in \citep{xiong2020layer}. Both models are constructed from 24 encoder layers and 12 decoder layers with 1024 hidden dimension and 4096 feedforward layer hidden dimension with 16 multi-head attention heads. For ZCode Dense, the number of parameters is 0.7B and vocab size is 250,000. The ZCode M3 has 32 experts, 5B parameters, and 128,000 vocab size. 

\noindent \textbf{Cluster Data and Training Settings} An in-house training set is used to offline train the clusters as indicated in Section \ref{sec:cluster_training}. This set encompasses 6 language directions. These directions are Spanish to English (EsEn), French to English (FrEn), Italian to English (ItEn), English to Spanish (EnEs), English to French (EnFr), and English to Italian (EnIt). To support multilinguality, we offline create for each language direction a number of clusters based on the hidden context vector similarity. Training the clusters for each language direction used 20 million sentence examples and was done using Faiss, where we run 20 training iterations of kmeans clustering. Faiss is an open source code for clustering training offered by Facebook \citep{johnson2019billion}.

\noindent \textbf{Validation Data} Table \ref{tab:input_test_set_sizes} indicates the number of sentence examples in our two in-house sets. EX means from English language pairs, while XE means to English language pairs.

\begin{table}[ht]
\caption{Statistics of the in-house validation sets. Set 1 is a mix of multiple domains with focus on news, while Set 2 is a mix of news, industry, government, and finance domains.} \label{tab:input_test_set_sizes}
\begin{tabularx}{\columnwidth}{?Y?Y?Y?Y? }
\toprule % for making lines bold
\specialrule{.1em}{.05em}{.05em} 
    
  \multicolumn{1}{ ?Y?}{\textbf{}}
& \multicolumn{1}{ Y?}{\textbf{Language Direction}}  
& \multicolumn{1}{ Y?}{\textbf{Set 1 Size}}
& \multicolumn{1}{ Y?}{\textbf{Set 2 Size}}
\\

\hline
\specialrule{.1em}{.05em}{.05em}
& EnEs & 5551 & 9639
\\
\cline{2-4}
\textbf{EX} & EnFr & 20038 & 13725 
\\\cline{2-4}
& EnIt & 9747 & 13747
\\\hline
\specialrule{.1em}{.05em}{.05em}
& FrEn & 20039 & 6149 
\\\cline{2-4}
\textbf{XE} & ItEn & 9747 & 7492 
\\\cline{2-4}
& EsEn & 5460 & 4539
\\\hline
\specialrule{.1em}{.05em}{.05em}
\end{tabularx}
\centering
\end{table}

\noindent \textbf{Inference Hardware and Environment} Inference experiments are carried out on a single NVIDIA Tesla V100 GPU. Float16 inference is enabled because it speeds up the inference while maintaining accuracy. Our inference environment is the highly optimized FasterTransformer from NVIDIA.

\subsection{Impact of Clustering on Vocab Projection Matrix Multiplication}
\label{sec:clus_matmul}

Our proposed method via clustering noticeably reduces the time to complete the vocab projection matrix multiplication. To verify this insight, we carried out an experiment to compare the proposed solution with the default inference without clustering on set 1. In ZCode Dense, the vocab projection runs in about 720 $\upmu$s, while the vocab projection runs in about 600 $\upmu$s for ZCode M3 as the vocab size is smaller. As shown in Table \ref{tab:tulgv1_moe_vocab_proj_impact_new}, the proposed method only activates up to 12.39\% on average out of the original weight matrix in ZCode Dense, while running 2.4x faster than the default matrix multiplication in beam 2, batch 20, and float16 inference. For example, the default elapsed time of the vocab projection step in ZCode Dense is 720 \textbf{$\upmu$s}, while the elapsed time using the proposed approach is 230.09 \textbf{$\upmu$s} for EnEs language direction. Along the same line, ZCode M3 activates up to 15.8\% of the original vocab projection weight matrix, while improving the time of vocab projection up to 2.6x.

\begin{table}[ht]
\caption{Proposed Method Gains of 2.4x to 2.6x on Vocab Projection using Set 1.}
\begin{tabularx}{\columnwidth}{?Y?Y?c?Y?c?}
\toprule % for making lines bold
\specialrule{.1em}{.05em}{.05em}

\multicolumn{1}{ ?Y?}{\textbf{Lang}}  
& \multicolumn{2}{ c?}{\textbf{ZCode Dense (720 \textbf{$\upmu$s} default)}}
& \multicolumn{2}{ c?}{\textbf{ZCode M3 (600 \textbf{$\upmu$s} default)}}
\\
\cline{2-5}
& \multicolumn{1}{ c?}{\textbf{Active Percentage (\%)}}
& \multicolumn{1}{ c?}{\textbf{Time \textbf{$\upmu$s}}}
& \multicolumn{1}{ c?}{\textbf{Active Percentage (\%)}}
& \multicolumn{1}{ c?}{\textbf{Time \textbf{$\upmu$s}}}
\\
\cline{1-5}
% Tulgv1
EnEs & 6.36 & 230.09 & 12.70 & 215.90
\\
\cline{1-5}
EnFr & 6.37 & 228.90 & 12.18 & 218.05 
\\
\cline{1-5}
EnIt & 7.46 & 231.84 & 15.80 & 227.80
\\
\cline{1-5}
FrEn & 16.50 & 348.49 & 15.70 & 230.50
\\
\cline{1-5}
ItEn & 10.05 & 274.90 & 13.60 & 217.30
\\
\cline{1-5}
EsEn & 12.39 & 299.97 & 12.70 & 213.50
% % XE average
% \textbf{XE AVG} & \textbf{xx} & \textbf{xx} & \textbf{xx} & \textbf{xx} & \textbf{xx} & \textbf{xx}
% \\
% \cline{1-7}
% % EX average
% \textbf{EX AVG} & \textbf{xx} & \textbf{xx} & \textbf{xx} & \textbf{xx} & \textbf{xx} & \textbf{xx}
\\
\hline
\specialrule{.1em}{.05em}{.05em}
\end{tabularx}
\label{tab:tulgv1_moe_vocab_proj_impact_new}
\centering
\end{table}

\subsection{End-to-end Accuracy and Speed Performance Comparison}
\label{sec:perf_eval}

Tables \ref{tab:clus_big_test_new} and \ref{tab:clus_auth_test_new} compare the speed percentage and BLEU scores of the ZCode Dense and ZCode M3 before and after clustering on our validation sets. Speed percentages are done based on the number of tokens per second in the default and proposed method, where higher numbers indicate more tokens per second for the proposed method. For ZCode Dense, end-to-end inference speed percentages range from 13.0\% to 25.7\%  across different language directions while maintaining the BLEU score and slightly increasing the memory required for the model due to the extra information needed by the clustering (see Table \ref{tab:clus_memory} for memory). Similarly, the proposed vocab projection method speeds up the ZCode M3 with a range between 6\% to 8.8\% across different language directions. It is worth noting that this performance evaluation is done in FasterTransformer, which showed noticeable speedups for encoder-decoder transformer architectures. In addition, authors in \citep{chen2018learning} have done extensive comparisons to the prior art and showed that clustering-based solutions are better than other solutions in bi-lingual settings on CPU. This paper extends \citep{chen2018learning}, scales the clustering to the multi-lingual setting as well as large vocab sizes on GPU, and shows positive results for both dense and sparsely activated transformers.

\begin{table}[ht]
\caption{Performance Evaluation of the Proposed Vocab Projection Method using Set 1.}
\begin{tabularx}{\columnwidth}{?c?Y?l?Y?c?l?Y?c? }
\toprule % for making lines bold
\specialrule{.1em}{.05em}{.05em} 

  \multicolumn{1}{ ?Y?}{\textbf{Model}}
& \multicolumn{1}{ Y?}{\textbf{Lang}}  
& \multicolumn{3}{ c?}{\textbf{beam=1, batch=1}}
& \multicolumn{3}{ c?}{\textbf{beam=2, batch=20}}
\\

\hline
\specialrule{.1em}{.05em}{.05em}
% \cline{3-8}

& \multicolumn{1}{ Y?}{\textbf{}}  
& \multicolumn{2}{ c?}{\textbf{BLEU}}
& \multicolumn{1}{ c?}{\textbf{Speedup (\%)}}
& \multicolumn{2}{ c?}{\textbf{BLEU}}
& \multicolumn{1}{ c?}{\textbf{Speedup (\%)}}
\\

\cline{3-8}
% \hline
% \specialrule{.1em}{.05em}{.05em}

&  & Baseline & Clus &  & Baseline & Clus & 
\\
\cline{2-8}
% Tulgv1
& EnEs & 43.29 & 43.21 & 25.70 & 43.89 & 43.81 & 13.00
\\
\cline{2-8}
& EnFr & 38.90 & 38.80 & 25.90 & 39.55 & 39.55 & 10.20
\\
\cline{2-8}
ZCode  & EnIt & 38.58 & 38.49 & 23.55 & 39.24 & 39.30 & 13.96 
\\
\cline{2-8}
Dense & FrEn & 44.83 & 44.80 & 22.25 & 45.37 & 45.46 & 15.51
\\
\cline{2-8}
& ItEn & 44.55 & 44.28 & 26.60 & 45.37 & 45.38 & 16.34 
\\
\cline{2-8}
& EsEn & 45.28 & 45.36 & 21.03 & 45.99 & 45.96 & 11.85
\\
\cline{2-8}
% XE average
& \textbf{EX AVG} & \textbf{39.50} & \textbf{39.40} & \textbf{25.00} & \textbf{40.14} & \textbf{40.15} & \textbf{11.70}
\\
\cline{2-8}
% EX average
& \textbf{XE AVG} & \textbf{45.03} & \textbf{44.80} & \textbf{23.30} & \textbf{45.57} & \textbf{45.60} & \textbf{14.60}
\\
\hline
\specialrule{.1em}{.05em}{.05em}

% % MoE
& EnEs & 45.99 & 45.96 & 6.80 & 46.07 & 46.09 & 7.87
\\
\cline{2-8}
& EnFr & 45.50 & 45.46 & 7.58 & 45.68 & 45.69 & 7.58
\\
\cline{2-8}
ZCode & EnIt & 43.09 & 43.08 & 6.59 & 43.31 & 43.35 & 8.84
\\
\cline{2-8}
M3 & FrEn & 50.06 & 49.92 & 4.70 & 50.21 & 50.31 & 4.90
\\
\cline{2-8}
& ItEn & 49.28 & 49.12 & 6.84 & 49.28 & 49.36 & 7.78
\\
\cline{2-8}
& EsEn & 50.01 & 50.04 & 7.50 & 50.21 & 50.32 & 6.00

\\
\cline{2-8}
% XE average
& \textbf{EX AVG} & \textbf{44.86} & \textbf{44.83} & \textbf{7.00} & \textbf{45.02} & \textbf{45.04} & \textbf{8.10}
\\
\cline{2-8}
% EX average
& \textbf{XE AVG} & \textbf{49.78} & \textbf{49.69} & \textbf{6.30} & \textbf{49.90} & \textbf{49.90} & \textbf{6.20}
\\
\hline
\specialrule{.1em}{.05em}{.05em}

\end{tabularx}
\label{tab:clus_big_test_new}
\centering
\end{table}

\begin{table}[ht]
\caption{Performance Evaluation of the Proposed Vocab Projection Method on Set 2 at beam=2, batch=20, and float16 inference.}
\begin{tabularx}{\columnwidth}{?Y?l?Y?c?l?Y?c? }
\toprule % for making lines bold
\specialrule{.1em}{.05em}{.05em} 

\multicolumn{1}{ ?Y?}{\textbf{Lang}}  
& \multicolumn{3}{ c?}{\textbf{ZCode Dense}}
& \multicolumn{3}{ c?}{\textbf{ZCode M3}}
\\

\cline{2-7}

& \multicolumn{2}{ c?}{\textbf{BLEU}}
& \multicolumn{1}{ c?}{\textbf{Speedup (\%)}}
& \multicolumn{2}{ c?}{\textbf{BLEU}}
& \multicolumn{1}{ c?}{\textbf{Speedup (\%)}}
\\

\cline{2-7}
& Baseline & Clus &  & Baseline & Clus & 
\\
\cline{1-7}
% Tulgv1
EnEs & 43.95 & 43.95 & 12.29 & 47.21 & 47.25 & 6.70
\\
\cline{1-7}
EnFr & 41.46  & 41.46 & 11.83 & 46.74 & 46.77 & 5.95
\\
\cline{1-7}
EnIt & 41.79 & 41.81 & 11.82 & 47.28 & 47.34 & 6.50 
\\
\cline{1-7}
FrEn & 46.31 & 46.24 & 11.49 & 48.64 & 48.75 & 6.00
\\
\cline{1-7}
ItEn & 44.15 & 44.01 & 13.34 & 47.89 & 47.99 & 6.48
\\
\cline{1-7}
EsEn & 45.02 & 44.89 & 12.05 &  47.56 & 47.68 & 7.00
\\
\cline{1-7}
% XE average
\textbf{EX AVG} & \textbf{42.40} & \textbf{42.40} & \textbf{12.00} & \textbf{47.07} & \textbf{47.12} & \textbf{6.40}
\\
\cline{1-7}
% EX average
\textbf{XE AVG} & \textbf{45.16} & \textbf{45.04} & \textbf{12.30} & \textbf{48.03} & \textbf{47.81} & \textbf{6.50}
\\
\hline
\specialrule{.1em}{.05em}{.05em}
\end{tabularx}
\label{tab:clus_auth_test_new}
\centering
\end{table}

\begin{table}[ht]
\caption{Memory cost Comparisons under Different Models at beam=2, batch=20 and float16 inference. Selected number of centroids vary a bit more for ZCode M3 than ZCode Dense.}
\begin{tabularx}{\columnwidth}{?Y?Y?Y?Y?Y? }
\toprule % for making lines bold
\specialrule{.1em}{.05em}{.05em} 

\multicolumn{1}{ ?Y?}{\textbf{}}  
& \multicolumn{2}{ c?}{\textbf{ZCode Dense}}
& \multicolumn{2}{ c?}{\textbf{ZCode M3}}
\\
\cline{2-5}
% \specialrule{.1em}{.05em}{.05em}

\textbf{Language Direction} & \textbf{Memory Baseline (GB)} & \textbf{Memory Clus (GB)} & \textbf{Memory Baseline (GB)} & \textbf{Memory Clus (GB)}
\\
\cline{1-5}
% % Tulgv1
EnEs & 4.37 & 4.39 & 10.19 & 10.31
\\
\cline{1-5}
EnFr & 4.37 & 4.39 & 10.19 & 10.35
\\
\cline{1-5}
EnIt & 4.37 & 4.39 & 10.19 & 10.29
\\
\cline{1-5}
FrEn & 4.37 & 4.39 & 10.19 & 10.35
\\
\cline{1-5}
ItEn & 4.37 & 4.39 & 10.19 & 10.28
\\
\cline{1-5}
EsEn & 4.37 & 4.39 & 10.19 & 10.37
\\
\hline
\specialrule{.1em}{.05em}{.05em}
\end{tabularx}
\label{tab:clus_memory}
\centering
\end{table}

\subsection{Human Evaluation}
\label{sec:human_eval}
We conduct 
quality evaluation to examine whether the proposed method retained the same output quality as Baseline and 
does not introduce any quality degradation.
For this investigation, we employ Direct Assessment human evaluation 
(DA; \citeauthor{bentivogli2018machine}, \citeyear{bentivogli2018machine}). 
In particular, we use segment-level contrastive source-based DA (contrastive DA; \citeauthor{akhbardeh2021findings}, \citeyear{akhbardeh2021findings}). 
In contrastive DA, human annotators see two randomly selected output segments generated 
by Baseline and our method side-by-side anonymized along with the corresponding source segment. 
Annotators are prompted to rate each output segment on a continuous scale of 0 to 100, 
based on the translation adequacy. 
Because both system output segments are presented simultaneously in contrastive DA, 
humans are encouraged to highlight translation differences with scores they assign, 
which allows us to capture quality differences between Baseline and our method effectively.
The contrastive DA score annotations were collected with the Appraise framework 
\citep{federmann2018appraise}.
We use paid professional annotators for all the evaluation tasks.

For evaluation data, we use 200,000 monolingual segments for each language direction sampled 
from our in-house evaluation data pool consisting of various data sources or domains. 
Among these test items, we confirmed that approximately 92\% of translations 
generated by our method and Baseline for each language direction are identical, 
which indicates that there was no quality degradation with these test items. 
Then we randomly sampled 1,700 output translation pairs among the non-identical pairs found  
in the remaining 8\% (roughly 16,000 pairs) for the contrastive DA evaluation.

Table~\ref{contrastiveDAResults} shows the contrastive DA results. 
We observed no cases where Baseline scores were better than those of our method with statistical significance. Moreover, our method outperformed Baseline for English into Spanish and French ($p<0.01$)
and English into Italian ($p<0.05$). 
These results clearly indicates that our method did not sacrifice any translation quality for the efficiency gains. 
Also, our method was better than Baseline with statistical significance
for all the from-English directions. 
This seems promising, but it still requires a deeper insight to justify overall quality improvement 
for from-English directions with these small positive deltas 
found in a limited number of language pairs.
Further investigation with a wider variety of language pairs is necessary 
to confirm such quality improvement exists.

\begin{table}[ht]
\caption{Contrastive DA Human Evaluation Results. Scores in the third and fourth columns are 
the mean of item scores for each language direction of each system.
Positive $\Delta$ indicates that the Clustering score is larger (better) than the Baseline score.  
We used Wilcoxon Rank Sum Test for significance testing. 
Score differences are statistically significant at $p<0.05$ ($\thinstar$), 
$p<0.01$ ($\thinstar\!\thinstar$), $p<0.001$ ($\thinstar\!\!\thinstar\!\!\thinstar$),
or not at all.} \label{contrastiveDAResults}
\begin{tabularx}{\columnwidth}{?Y?Y?Y?Y?Y?Y? }
\toprule % for making lines bold
\specialrule{.1em}{.05em}{.05em} 
    
  \multicolumn{1}{ ?Y?}{\textbf{Model}}
& \multicolumn{1}{ Y?}{\textbf{Language Direction}}  
& \multicolumn{1}{ Y?}{\textbf{Baseline}}
& \multicolumn{1}{ Y?}{\textbf{Clustering}}
& \multicolumn{1}{ Y?}{$\boldsymbol{\Delta}$}
& \multicolumn{1}{ Y?}{\textbf{\textit{P}-value}}
\\

\hline
\specialrule{.1em}{.05em}{.05em}
& EnEs & 87.20 & 88.10 & 0.90 & $\thinstar\!\thinstar$
\\
\cline{2-6}
& EnFr & 88.10 & 88.90 & 0.80 & $\thinstar\!\thinstar$
\\\cline{2-6}
ZCode & EnIt & 85.10 & 85.5 & 0.40 & $\thinstar$
\\\cline{2-6}
M3 & FrEn & 86.70 & 87.1 & 0.40 &  
\\\cline{2-6}
& ItEn & 79.40 & 79.40 & 0.00 &
\\\cline{2-6}
& EsEn & 86.10 & 86.60 & 0.50 &
\\\hline
\specialrule{.1em}{.05em}{.05em}
\end{tabularx}
\label{tab:vocab_proj_human_eval}
\centering
\end{table}

\subsection{Clustering Hyperparameters Selection}
\label{sec:clus_centroids_topk}

To demystify the impact of the number of K likely tokens and centroids in the proposed method, we carried out two experiments on ItEn language direction on a validation set. The results of these experiments were also consistent for other language pairs.

\begin{figure} [htbp]
      \centering
             \includegraphics[width=1\linewidth]{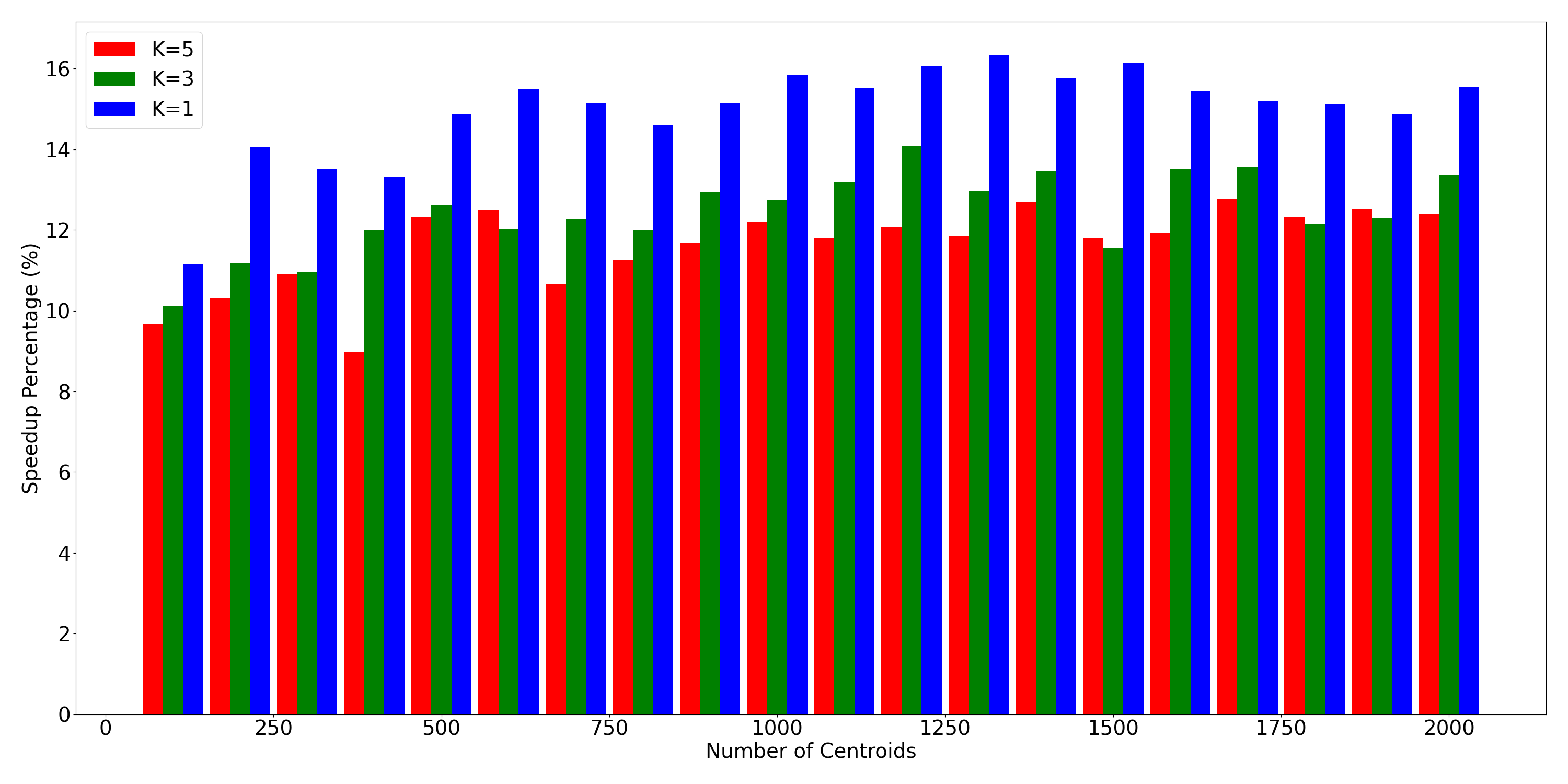}
      \caption{Impact of Varying the Number of Centroids and K likely tokens on Speed Percentage carried out on ItEn language direction from Development Set.}
      \label{fig:clusters_topK}
\end{figure}

The goal of the first experiment is to select the best number of centroids and K likely tokens for ItEn, where we varied the number of centroids from 100 to 2000 with a step size of 100 and setting the K likely tokens to either 1, 3, or 5. Figure \ref{fig:clusters_topK} shows the outcome of this experiment for the ItEn language direction on ZCode Dense, where speed percentages are calculated to the ZCode Dense without clustering. From the figure, we can observe that the highest speed percentages usually occur at the higher centroids. In this case, 1300 centroids is the best setting in terms of speed. In addition, we can observe that K=1 leads to the highest speed percentages among K=3 and K=5. This observation goes with one's intuition as K=1 uses only the most likely predicted token to build the activated weight vocab indices for each cluster.

\begin{figure} [htbp]
      \centering
             \includegraphics[width=1\linewidth]{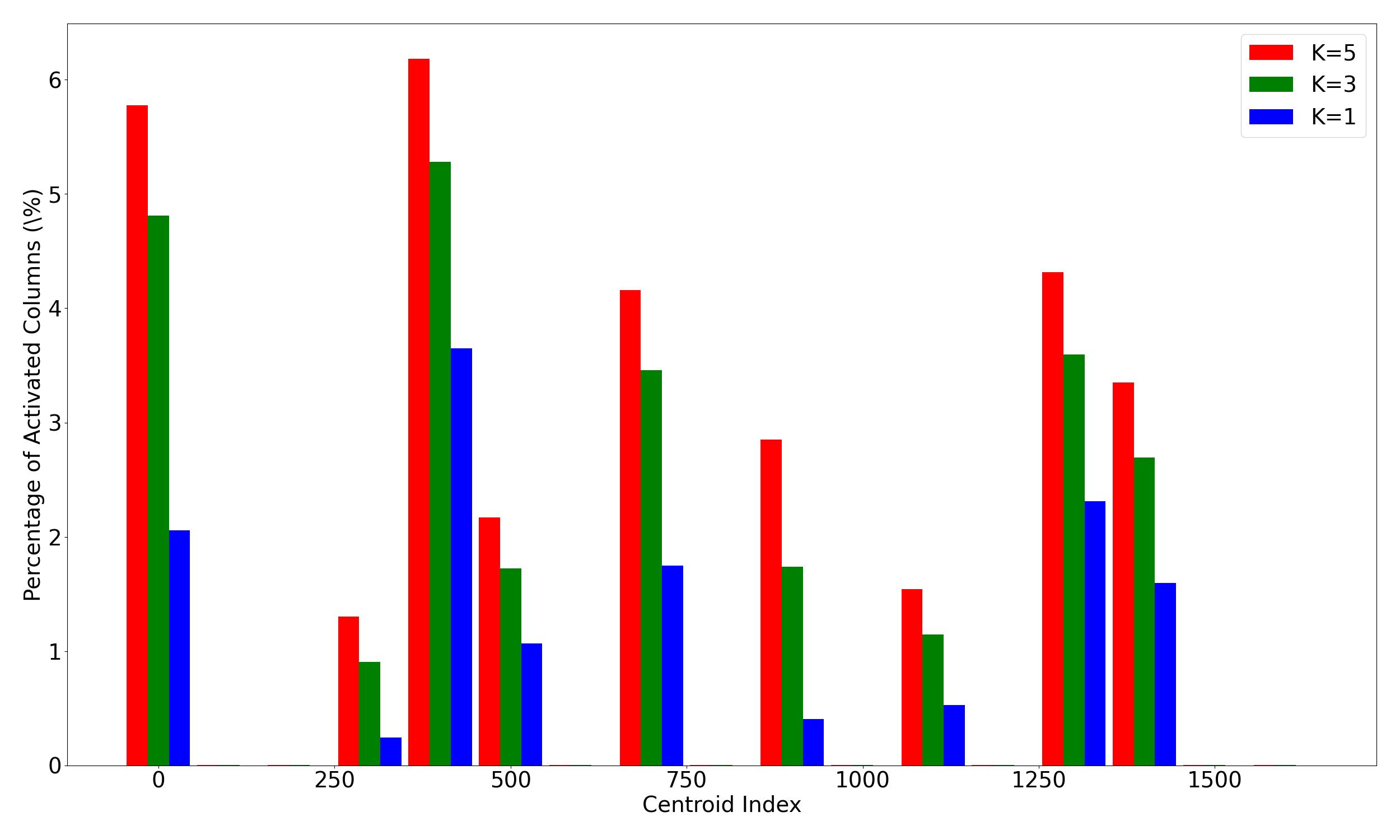}
      \caption{Comparison between K=5, K=3, and K=1 likely tokens in terms of Percentage Activated Weight Columns out of Vocab Elements When Number of Centroids = 1300 for ItEn.}
      \label{fig:topK1_topK3_topk5_active}
\end{figure}

To confirm this intuition, we carried out another experiment for ItEn under ZCode Dense while setting the number of centroids to 1300. In this experiment, we recorded the percentage of weight active columns at each centroid index from 0 to 1299 for each of K=5, K=3, and K=1. As shown in Figure \ref{fig:topK1_topK3_topk5_active},  K=1 has less activated weight indices for each centroid index relative to K=3 and K=5. If a hidden context vector is a member of cluster index 0 for example, K=5, K=3, and K=1 have around 6\%, 5\%, and 2\% of weight matrix active, respectively. These percentages confirm that K=1 usually leads to the highest speed percentages among K=3 and K=5.

% Can you increase the font size a bit in the figure?

%%%%%%%%%%%%%%%%%%%%%%%%%%

\section{Conclusion}
\label{conclusion}

This paper proposes a fast vocab projection method via clustering for multilingual neural machine translation with large vocab sizes and practically applied models on GPU. The method splits the large vocab search space into smaller subspaces to run efficient GPU inference. Results reveal end-to-end speed improvements up to 25\% while maintaining the BLEU, and up to 2.6x speed improvement on the vocab projection step. We conducted human evaluations to verify the translation quality. In the future, we wish to explore our method in an X-Y machine translation scenario. %\footnote{{\url{https://www.microsoft.com/en-us/research/blog/microsoft-translator-enhanced-with-z-code-mixture-of-experts-models/}}}

%%%%%%%%%%%%%%%%%%%%%%%%%%

%\clearpage

\small

\bibliographystyle{apalike}
\bibliography{amta2022}

\end{document}